\documentclass{article}

\usepackage[preprint]{neurips_2025}
\usepackage[utf8]{inputenc} % allow utf-8 input
\usepackage[T1]{fontenc}    % use 8-bit T1 fonts
\usepackage{hyperref}       % hyperlinks
\usepackage{url}            % simple URL typesetting
\usepackage{booktabs}       % professional-quality tables
\usepackage{amsfonts}       % blackboard math symbols
\usepackage{nicefrac}       % compact symbols for 1/2, etc.
\usepackage{microtype}      % microtypography
\usepackage{xcolor}         % colors
\usepackage{amsmath}
\usepackage{graphicx} 
\usepackage{pdfpages}
\usepackage{float}
\usepackage{subcaption}
\usepackage{caption}
\usepackage{lipsum}

\begin{document}
% Note. For the workshop paper template, both \title{} and \workshoptitle{} are required, with the former indicating the paper title shown in the title and the latter indicating the workshop title displayed in the footnote. 
\title{Modular Diffusion Policy Training: Decoupling and Recombining Guidance and Diffusion for Offline RL}
\author{%
  Zhaoyang Chen \\
  Department of Mechanical Engineering\\
  Iowa State University\\
    2529 Union Drive
    Ames, IA 50011-2030 \\
  \texttt{zchen1@iastate.edu} \\
  % examples of more authors
  \And
  Cody Fleming \\
  Department of Mechanical Engineering \\
  Iowa State University\\
    2529 Union Drive
    Ames, IA 50011-2030 \\
  \texttt{flemingc@iastate.edu} \\}

% The \author macro works with any number of authors. There are two commands
% used to separate the names and addresses of multiple authors: \And and \AND.
%
% Using \And between authors leaves it to LaTeX to determine where to break the
% lines. Using \AND forces a line break at that point. So, if LaTeX puts 3 of 4
% authors names on the first line, and the last on the second line, try using
% \AND instead of \And before the third author name.

% \author{%
%   David S.~Hippocampus\thanks{Use footnote for providing further information
%     about author (webpage, alternative address)---\emph{not} for acknowledging
%     funding agencies.} \\
%   Department of Computer Science\\
%   Cranberry-Lemon University\\
%   Pittsburgh, PA 15213 \\
%   \texttt{hippo@cs.cranberry-lemon.edu} \\
%   % examples of more authors
  % \And
  % Coauthor \\
  % Affiliation \\
  % Address \\
  % \texttt{email} \\
  % \AND
  % Coauthor \\
  % Affiliation \\
  % Address \\
  % \texttt{email} \\
  % \And
  % Coauthor \\
  % Affiliation \\
  % Address \\
  % \texttt{email} \\
  % \And
  % Coauthor \\
  % Affiliation \\
  % Address \\
  % \texttt{email} \\
% }

\maketitle

\begin{abstract}
Classifier-free guidance has shown strong potential in diffusion-based reinforcement learning. However, existing methods rely on joint training of the guidance module and the diffusion model, which can be suboptimal during the early stages when the guidance is inaccurate and provides noisy learning signals. In offline RL, guidance depends solely on offline data—observations, actions, and rewards—and is independent of the policy module’s behavior, suggesting that joint training is not required. This paper proposes modular training methods that decouple the guidance module from the diffusion model, based on three key findings:

Guidance Necessity: We explore how the effectiveness of guidance varies with the training stage and algorithm choice, uncovering the roles of guidance and diffusion. A lack of good guidance in the early stage presents an opportunity for optimization.

Guidance-First Diffusion Training: We introduce a method where the guidance module is first trained independently as a value estimator, then frozen to guide the diffusion model using classifier-free reward guidance. This modularization reduces memory usage, improves computational efficiency, and enhances both sample efficiency and final performance.

Cross-Module Transferability: Applying two independently trained guidance models—one during training and the other during inference—can significantly reduce normalized score variance (e.g., reducing IQR by 86\%). We show that guidance modules trained with one algorithm (e.g., IDQL) can be directly reused with another (e.g., DQL), with no additional training required, demonstrating baseline-level performance as well as strong modularity and transferability.

We provide theoretical justification and empirical validation on bullet-D4RL benchmarks. Our findings suggest a new paradigm for offline RL: modular, reusable, and composable training pipelines.
\end{abstract}

\section{Introduction}
%Note: This work has been submitted for blind review. Please do not cite or promote it publicly until the review process is complete.
Diffusion models have recently emerged as a promising approach for sequential decision-making, particularly in offline reinforcement learning, where policies must be learned entirely from static datasets without any interaction with the environment. By formulating policy learning as a conditional generative process, diffusion-based policies are capable of modeling complex, multimodal behaviors and have demonstrated strong empirical performance across a wide range of continuous control and robotic manipulation tasks \citep{janner2022planning,chi2023diffusion,wang2023diffusion}
. In many cases, diffusion policies outperform traditional reinforcement learning methods, as highlighted in prior work \citep{hansen2023idql, chen2024dtql}.

A central challenge in diffusion-based policy learning lies in injecting reward information into the denoising process, especially during fine-tuning via reinforcement learning. One of the most widely used methods is Classifier-Free Guidance (CFG), which biases sampling toward high-reward trajectories \citep{ho2022classifierfree}.

A clear issue arises when the guidance module and the diffusion model are trained jointly. The role of the guidance module is to steer the diffusion process toward generating actions that lead to higher rewards. However, before the guidance module has converged, it cannot provide effective guidance. As a result, during the early stages of training, the diffusion model receives little to no meaningful feedback from the guidance module, which may even misguide the diffusion process and degrade its performance~\citep{kim2023refining}.

\citep{zhang2023towards} and other mainstream existing methods \citep{uehara2025inference,kim2023refining} solve this problem by pretraining the diffusion models without the guidance module. However, since improper guidance can jeopardize diffusion training, instead of removing the guidance, our paper solves this problem in the opposite way, pretraining the guidance module before training the diffusion model. More fundamentally, the guidance module is trained using observations, actions, and rewards from a dataset, data that are independent of the actions generated by the diffusion model. Therefore, the guidance module can be trained separately using supervised learning. Once trained, it can then be used to effectively guide the diffusion model. Importantly, this separation does not reduce exploration, as offline reinforcement learning relies entirely on fixed datasets, without any online exploration. Additionally, separate training reduces the peak memory required during training, even if the overall computation remains the same. This makes the approach more efficient and practical \citep{kim2023refining}.

Another advantage of training the guidance module separately is its flexibility. There is not just a single way to modularize the model components. One known issue with classifier-free guidance (CFG) is that the guidance module can produce inaccurate outputs: as noted in ~\cite{vanHasselt2010doubleq}, reward estimation inherently involves both overestimation and underestimation. Our Double Guidance diffusion architecture addresses this by using a differently seeded version of the guidance network during inference, distinct from the one used in training. This replacement helps correct inaccurate estimations and significantly reduces variance.

Since different runs of the modules are interchangeable, we further investigated whether different model architectures could be used interchangeably in a plug-and-play manner. To explore this flexibility, we cross-combined components from models trained under different frameworks. Specifically, we took the guidance module from a jointly trained IDQL model (i.e., the IDQL classifier trained alongside the IDQL diffusion model) and paired it with a completely separate DQL diffusion model—one that had never been trained in conjunction with the IDQL classifier. Despite the fact that these two components had never interacted during training, the resulting hybrid model was functional.

Surprisingly, this forcibly combined model—formed by merging the IDQL guidance module with the DQL diffusion model at each saved checkpoint—outperformed both the standalone IDQL and DQL models. We saved checkpoints every 400 training steps and evaluated the hybrid model accordingly. Although the combination was applied solely during inference, the performance curve of the hybrid model rose more rapidly than those of either original model. Ultimately, it matched the performance baseline of DQL, which was the better of the two.

This result strongly suggests that the relationship between the guidance module and the diffusion model is far more modular and flexible than previously assumed. Such modularity not only offers practical benefits but also highlights a promising direction for future research~\cite{zhao2025ctrl}.

To solidify our empirical findings, we provide theoretical justification for why a pretrained guidance model can accelerate training and why the modules are interchangeable across different diffusion models. This theoretical foundation explains the effectiveness of the offline RL models we implemented and evaluated, showing them as concrete instances of this broader principle. %More generally, this method has far-reaching applications beyond the specific cases explored in our work.

\section{Preliminaries}

\subsection{Q-Guided Diffusion}
In offline RL, the goal is to learn a policy $\pi_\theta(a|s)$ that maximizes expected returns given a fixed dataset $\mathcal{D} = \{s, a, r, s'\}$. When using diffusion models as the policy, offline reinforcement learning methods, including DQL~\cite{dql2022planning}, EDP~\cite{edp2024kang}, and IQL~\cite{idql2023hansen}, all employ a similar conditional denoising diffusion process to parameterize the policy $\pi_\theta(a|s)$. Specifically, they adopt the forward noising process:
\begin{equation}
a^k = \sqrt{\bar{\alpha}_k} a^0 + \sqrt{1 - \bar{\alpha}_k} \cdot \epsilon, \quad \epsilon \sim \mathcal{N}(0, I)
\end{equation}
and learn a denoising network $\epsilon_\theta(a^k, k, s)$ to recover the clean action $a^0$. During inference or training, the clean action is reconstructed via running the full reverse denoising process as in standard DDPM sampling(DQL, IDQL) or one-step (EDP) approximation:
\begin{equation}
\hat{a}^0 = \frac{1}{\sqrt{\bar{\alpha}_k}} a^k - \frac{\sqrt{1 - \bar{\alpha}_k}}{\sqrt{\bar{\alpha}_k}} \cdot \epsilon_\theta(a^k, k, s)
\end{equation}
The diffusion loss for the methods below is 
\begin{align}
\mathcal{L}_{\text{diff}} &= \mathbb{E}_{a_0, \epsilon, t} \left[ \left\| \epsilon_\theta(a_t, t, c) - \epsilon \right\|^2 \right],
\mathcal{L}_Q = - \frac{\mathbb{E}[Q(s, a)]}{\left| \mathbb{E}[Q'(s, a)] \right|}, 
\mathcal{L}_{\text{actor}} = \mathcal{L}_{\text{BC}} + \eta \cdot \mathcal{L}_Q, \quad
\end{align}
As shown above, while the denoising loss trains the diffusion model $\epsilon_\theta(a^k, k, s)$ to match the offline behavior distribution, Q-guided diffusion with classifier-free guidance includes a loss term that steers the denoising process toward high-value actions. During training, the Q-network $Q_\phi(s, a)$ is optimized via temporal-difference learning to estimate the expected return of actions.

At inference time, the diffusion model generates actions by sampling from noise and iteratively denoising, adjusting the sampling process using the gradient of the Q-function:
\[
\tilde{a}^0 = a^0 + \lambda \nabla_a Q_\phi(s, a^0)
\]
where $\lambda$ is a scaling factor controlling the strength of the guidance. 
\subsection{Variations of Q-guided diffusion}
\textbf{Diffusion Q-learning (DQL)}: The DQL model modified the guidance module and applied a variation of the Deep Q network called double Q, which is a common procedure for stabilizing Q value estimation ~\cite{vanHasselt2010doubleq}.
\begin{equation}
\mathcal{L}_\pi^{\text{DQL}}(\theta) = - \mathbb{E}_{s \sim \mathcal{D}} \left[ \min \left( Q_{\phi_1}(s, a^0), Q_{\phi_2}(s, a^0) \right) \right], \quad \text{where } a^0 \sim \pi_\theta(\cdot | s)
\end{equation}

\textbf{Implicit Diffusion Q-learning (IDQL)} represents a different class of Q value estimator based on \textit{likelihood-weighted policy learning}. IQL maintains a stochastic policy $\pi_\theta(a|s)$, but instead of sampling new actions during training, it optimizes the policy via weighted behavior cloning over dataset actions:
\begin{equation}
\max_\theta \mathbb{E}_{(s, a) \sim \mathcal{D}} \left[ f(Q_\phi(s,a)) \cdot \log \pi_\theta(a|s) \right]
\end{equation}
where $f(Q_\phi(s,a))$ is an advantage-based weight. This approach avoids out-of-distribution actions, but requires that $\pi_\theta$ be a distribution with \textit{tractable log-likelihood}. The loss of it is expressed as 
\begin{equation}
\mathcal{L}_\pi^{\text{IQL}}(\theta) = - \mathbb{E}_{(s, a) \sim \mathcal{D}} \left[ w(s, a) \cdot \log \pi_\theta(a | s) \right]
\end{equation}
where $w(s,a)$ is an advantage-based weight derived from the learned Q-function. 
Notice that this IDQL baseline only does behavior cloning, with no $L_Q$ term in the diffusion training. It has a Q-based guidance term trained, but is only trained with supervised learning and applied only in the inference step. The success of IDQL inspired the decoupling methods proposed in this paper.

\textbf{Efficient Diffusion Policy (EDP)} introduces a key modification: \textit{one-step denoising}. Instead of running the full reverse chain to generate actions, EDP corrupts a dataset action $a^0$ to $a^k$, and then reconstructs an approximate clean action $\hat{a}^0$ using a denoise backward pass:
\begin{equation}
\hat{a}^0 = \frac{1}{\sqrt{\bar{\alpha}_k}} a^k - \frac{\sqrt{1 - \bar{\alpha}_k}}{\sqrt{\bar{\alpha}_k}} \cdot \epsilon_\theta(a^k, k, s)
\end{equation}
This \textit{action approximation} replaces expensive sampling with a lightweight inference step, making EDP orders of magnitude faster to train. This one-step denoising relies more on the guidance module, which will be mentioned in the section \ref{sec:Experiments}.

These three methods are the baselines of our methods and are Temporal-Difference(TD)-based methods. Our modifications are typically only applied to TD-based methods and cannot be applied to Trajectory-Based Reward guidance. Trajectory-based diffusion models (e.g., Diffuser~\cite{janner2022planning}) use return annotations for full sequences $\tau = (s_0, a_0, ..., s_T)$. However, experiments showed low convergence rates and suboptimal precisions in their Q-value predictions, and therefore, Guidance-First Diffusion Training showed little improvement. The reason for this observation is supposed to be: when the trajectory is long, the return may be noisy due to stochastic behavior policies, so reward attribution over entire sequences can be ambiguous.  In contrast, TD-based methods that work on $(s, a)$ pairs avoid this issue and support more granular, local learning signals.

\section{Methodology}
% There are two folds in separating reward estimation from policy learning diffusion:
% Guidance-First Diffusion Training (GFDT) and modular composition of network modules without co-training.

% Our proposed method,GFDT, addresses the instability of co-training guidance and policy by fully decoupling their training processes. First, we train a \emph{Learned Guidance Module (LGM)} on the offline dataset. This module can be a reward estimator, and it is trained via standard supervised learning to estimate $r(s, a)$. Since the dataset is fixed and does not depend on the current policy, the LGM can be trained independently without interaction with the environment.
% Once trained, the guidance module is frozen and used to guide the diffusion model. The diffusion model learns to generate actions conditioned on state by optimizing a policy gradient objective, where the reward signal is computed using the fixed LGM. This leads to more efficient training, as the guidance signal is consistent and no longer evolves with the policy. Compared to previous approaches that jointly train both components, GFDT prevents early-stage noisy gradients from under-trained guidance models, avoids feedback loops, and ensures more meaningful alignment between reward estimation and policy updates.

% We propose a modular composition of the guidance module and the diffusion model, where the two components do not need to be trained jointly, contrary to the approach in~\cite{ho2022classifierfree}. 
Having introduced the GFDT method and modular module composition in the introduction, we now extend the discussion with a mathematical model to justify the effectiveness of our modular training approach. First, we explain why a pretrained, converged model can provide more accurate guidance to the diffusion model compared to an unconverged model that is trained jointly with it, thereby accelerating the training process. Next, we prove that as long as the guidance model can provide a direction that leads to reward improvement and the step size is small enough, it is sufficient for guiding the diffusion model, even if it was not co-trained with it.
\subsection{Theoretical Motivation: Why Guidance Improves Diffusion Policies}

We build on the theoretical foundation established by \citep{fujimoto2019benchmarking}, which formalizes convergence guarantees for policy learning from fixed offline datasets. Under the assumption of a deterministic MDP and a coherent batch $\mathcal{B}$--meaning that, for all $(s,a,s') \in \mathcal{B}$, $s' \in \mathcal{B}$ unless $s'$ is terminal--the proposed Batch-Constrained Q-Learning (BCQL) algorithm is proven to converge to an optimal batch-constrained policy $\pi^*$.

\paragraph{Theorem 1 \citep{fujimoto2019benchmarking}}  
\textit{Given a deterministic MDP and coherent batch $\mathcal{B}$, along with standard Robbins-Monro convergence conditions on the learning rate, BCQL converges to $Q^{\pi_\mathcal{B}}(s, a)$, where $\pi^*(s) = \arg\max_{a \text{ s.t. } (s,a) \in \mathcal{B}} Q^{\pi_\mathcal{B}}(s, a)$. This policy is guaranteed to match or outperform any behavioral policy contained in the dataset.}

This result motivates our design choice: pre-training a guidance policy on offline data before conditioning a diffusion model on it. The batch-constrained approach ensures that guidance improves progressively as long as it remains within the support of the offline dataset, avoiding the instability often associated with end-to-end training on noisy value estimates. We hypothesize that high-quality guidance—either through behavioral priors or value-based corrections—significantly improves this process.

In practice, this results in a gradient update of the form:

\begin{equation}
\nabla_\theta \mathcal{L}_{\text{total}} = \nabla_\theta \mathbb{E}[Q(s,a)] - \lambda \nabla_\theta \mathcal{L}_{\text{BC}},
\end{equation}

where $\lambda$ controls the strength of the prior. To further bias sampling toward high-value actions, we incorporate value function gradients into the reverse diffusion steps. Specifically, we modify the noise prediction as:
\begin{equation}
\epsilon_\theta(a_t, s, t) \leftarrow \epsilon_\theta(a_t, s, t) + \alpha \nabla_{a_t} Q(s, a_t),
\end{equation}
where $\alpha$ is a guidance coefficient. This effectively performs value-aware perturbations during generation and encourages $a_0$ to align with the high-reward manifold. If $Q(s,a)$ is accurate within the support of $\mathcal{D}$, this can be seen as a form of approximate gradient ascent toward $\arg\max_a Q(s,a)$, while remaining within the data distribution. Empirically, we observe that removing either guidance degrades performance, consistent with our theoretical intuition.
\subsection{Theoretical Insight: Why Independent Value Guidance Works}

This subsection proves our second proposition: not only joint training of the guidance and the diffuser unnecessary, but also the value Q-network can be modularized (plug-and-play) and directly applied to a diffusion network, as long as it is a precise estimator of the value of the states. Although our Q-network $Q_\phi$ is trained independently of the diffusion model, we now sketch why its gradient can still reliably guide sampling toward higher-reward regions.

We consider a diffusion-based sampling process perturbed by the gradient of $Q_\phi$:
\begin{equation}
a_{t+1} = a_t + \eta_t \cdot \frac{\nabla_a Q_\phi(a_t)}{\|\nabla_a Q_\phi(a_t)\|} + \sqrt{2\tau_t}\xi_t
\end{equation}

Here, $\eta_t$ is a small step size, and $\tau_t$ controls the annealed noise. Crucially, the direction $\nabla_a Q_\phi(a_t)$ provides a reward-sensitive vector field.

\noindent\textbf{Observation:} \textit{As long as this gradient is positively correlated with the ideal sampling direction (i.e., it roughly points toward higher-density or higher-reward regions), even a pretrained $Q_\phi$ can iteratively bias the sample path toward improved quality.}

\noindent This follows from the observation that if $\theta_t = \angle(\nabla_x Q_\phi, \nabla_x \log p(x))$ and $\cos\theta_t > 0$, then over many steps:
\begin{equation}
\mathbb{E}[Q_\phi(x_{t+1}) - Q_\phi(x_t)] \approx \eta_t \cdot \cos\theta_t \cdot \|\nabla_x Q_\phi\| + \text{(noise)}
\end{equation}
\noindent Therefore, for sufficiently small $\eta_t$, the guided updates are directionally aligned with the increase of $Q_\phi$, even if $Q_\phi$ was trained separately. Since noise is Gaussian noise,\begin{equation}
\lim_{T \to \infty} \frac{1}{T} \sum_{t=1}^{T} \xi_t = 0 \quad \text{(in expectation)}
\end{equation}
Therefore, with a sufficiently small step size and a sufficient number of steps, the cumulative noise tends to zero in expectation, allowing the function to converge and stop at the optimal point rather than overshooting it. The algorithm is less likely to get trapped in local minima, as the injected noise enables exploration beyond narrow local optima.

\noindent\textbf{Conclusion:} \textit{Even without joint training, a pretrained Q-function can steer the diffusion process toward high-reward samples via its gradient, provided the guidance direction maintains positive alignment over time.}

\section{Experiments}\label{sec:Experiments}
\subsection{Datasets and Evaluation Protocol}
We evaluate our method on 8 standard tasks from the PyBullet D4RL benchmark~\citep{fu2020d4rl}. To ensure fair comparison, we re-train three seeds of three representative diffusion-based offline RL algorithms—EDP, DQL, and IDQL—based on publicly available code from Dong et al.~\citep{dong2024cleandiffuser}, and evaluate them in the PyBullet-based Gymnasium environments~\citep{farama2024gymnasium}. The application of our framework is not limited to these methods; rather, these methods serve as illustrative examples of our theory, which can be easily extended to more diffusion methods. All models are trained using the D4RLMuJoCoTD Dataset~\citep{fu2021d4rl}. It provides pre-collected trajectories of varying quality, including expert, medium-expert, medium-replay, and medium datasets, enabling rigorous training of offline RL algorithms under diverse data distributions. The evaluation is done in randomly initialized environments. All the gradient steps mentioned are with respect to a batch size of 256.

Wherever possible, we adopt the original hyperparameter settings from the paper of ~\citep{dong2024cleandiffuser}. Importantly, we do not modify any training-related components—including the optimizer, learning rate, batch size, architecture, or loss function. This design choice is intentional: a key strength of our method is that it achieves superior performance without requiring any changes to the original training recipe. This highlights the robustness and plug-and-play nature of our approach. Final results are reported in Table~\ref{tab:guidance_results}, averaged over 50 evaluation episodes. Performance is measured by normalized return, and we report both the mean and variance.

\subsection{The Role of the Guidance Network}

All experiments were conducted using PyTorch on a high-performance computing (HPC) cluster. We adhered to the assigned framework for our work because the focus of this paper is not directly related to computational power. The computational cost remains unchanged; only the order of computation was modified, or the training of the guidance module is entirely replaced by loading checkpoints. 

When training diffusion-based reinforcement learning models, reward signals are typically seen as essential for learning useful behaviors. To analyze the role of reward guidance and whether the guidance is removable or replaceable, we conducted an ablation study of a series of training sessions. TD-based diffusion models, DQL and EDP, with noise as guidance, without guidance, with baseline guidance, with Pretrained guidance, and with Pretrained guidance but continue training it when training the diffusion. To get a fair comparison, we saved a checkpoint every 400 gradient steps during training and evaluated them under randomized environments to measure the reward.

\begin{figure}[htbp]
\centering

\begin{subfigure}[b]{0.45\textwidth}
\includegraphics[width=\linewidth, trim=0 0 0 0, clip]{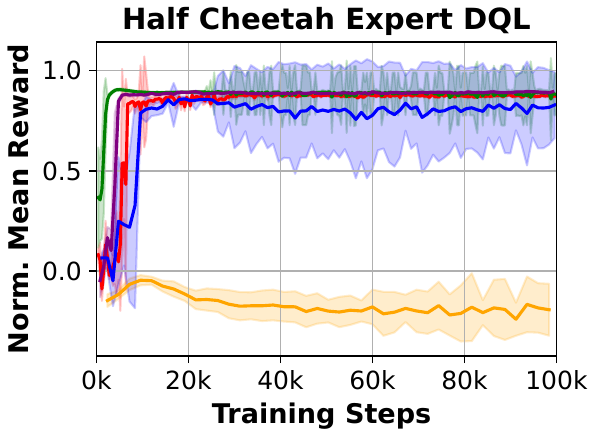}
\caption{Half Cheetah Expert DQL}
\label{fig:HalfCheetahExpertDQL}
\end{subfigure}
\hfill
\begin{subfigure}[b]{0.45\textwidth}
\includegraphics[width=\linewidth, trim=0 0 0 0, clip]{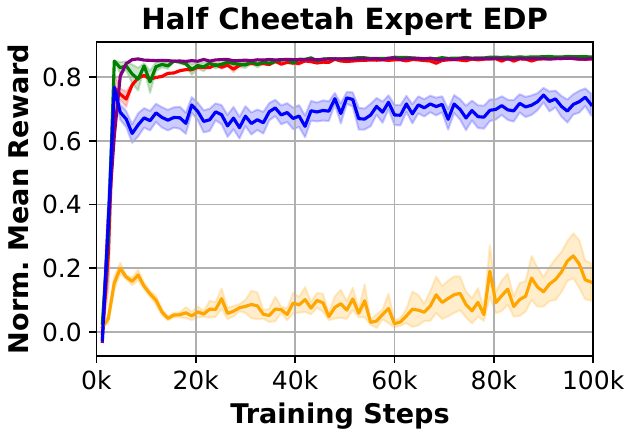}
\caption{Half Cheetah Expert EDP}
\label{fig:HalfCheetahExpertEDP}
\end{subfigure}

\vspace{0.8em}

\includegraphics[width=\linewidth, trim=0 0 0 0, clip]{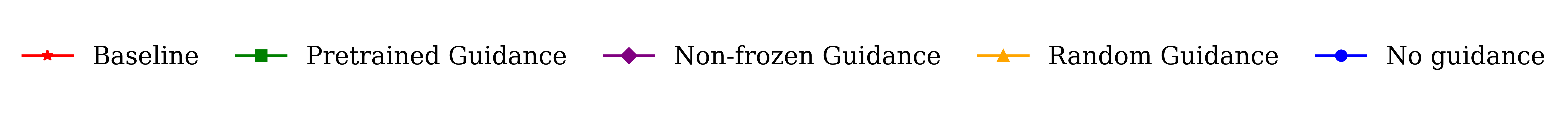}

\caption{Comparison between Half Cheetah Expert DQL and EDP with various guidance signals.}
\label{fig:HalfCheetahComparison}
\end{figure}

% Right figure: legend
% \begin{subfigure}[b]{0.31\textwidth}
% \includegraphics[width=\linewidth, trim=20 20 20 20, clip]{figures/legend_full.png}
% \caption*{Legend}
% \end{subfigure}

% \caption{Comparison between Half Cheetah Expert DQL and EDP with various guidance signals.}
% \label{fig:HalfCheetahComparison}
% \end{figure}

These results reveal that guidance is not uniformly essential--its importance is modulated by the stage of training and the algorithm itself. The improvement of no-guidance diffusion in the early stages is faster than that of the baseline, showing that, in the early stage, being guided by a guidance module that has not converged can jeopardize training. This result is further validated by the fact that when the guidance signal consisted purely of noise, it completely disrupted the training process.

However, precise guidance is necessary: those with guidance exhibited substantially lower variance in reward compared to the no-guidance group, indicating that guidance reduced variations in the learning dynamics. Removing guidance also leads to an initial high-performance peak to near baseline level, followed by a drop in test returns—a typical overfitting phenomenon. Diffusion training thus benefits from early stopping. EDP relies more on the guidance module than DQL, showing the importance of guidance to the one-step DPM-Solver.

Since misguidance can jeopardize training, what about ``good'' guidance? Pretrained guidance accelerates learning: We also conducted another model training on the guidance module that was pretrained before the training of the diffusion model, and then frozen in the diffusion training stage. In all such configurations, models demonstrated faster increases in reward. This supports our hypothesis that a well-trained guidance module can significantly accelerate diffusion training. If the diffusion model begins training before the guidance is properly learned, it may adapt to incorrect strategies, resulting in wasted training effort in the early training stage.

Effectiveness under few-shot scenarios: The benefit of pretrained guidance was particularly evident in few-shot settings, where training resources are limited. The GFDT always has better peak performance.These findings validate our theoretical assumption that guidance, when properly trained ahead of time, can direct the learning process of diffusion models more efficiently and reliably. Guidance consistently improved training efficiency across MuJoCo tasks. 

Another important point to note is that the guidance module must be locked. This is because, according to our previous experiments, if the guidance module is not locked but instead also opened up for further training, the advantage becomes significantly reduced. In other words, this limitation is mainly caused by overfitting resulting from excessive training of the guidance module.
Key metrics:

\begin{table}[ht]
\centering
\caption{Performance comparison between GFDT and baseline across selected D4RL MuJoCo Bullet tasks. All differences marked with \dag~are statistically significant (Mann-Whitney U test, $p < 0.0001$, large effect size). Variance heterogeneity is assessed via Levene's test.}
\label{tab:guidance_results}
\setlength{\tabcolsep}{3.5pt} % 默认是6pt
\begin{tabular}{lcccc}
\toprule
\textbf{Env.} & \textbf{Peak (\%)} & \textbf{AUC (\%)} & \textbf{Early} & \textbf{Levene (F/p)} \\
\midrule
HalfCheetah-M\dag       & 11.88 & 25.24 & 0.2457 & 13.43/0.0003 \\
HalfCheetah-M-E\dag     & 2.15  & 8.69  & 0.4982 & 23.91/0.0000 \\
HalfCheetah-E\dag       & 2.51  & 6.98  & 0.4479 & 11.36/0.0008 \\
Walker2d-M\dag          & 0.27  & 8.57  & 0.4327 & 13.49/0.0003 \\
Walker2d-M-E\dag        & 4.47  & 77.65 & 0.3062 & 187.00/0.0000 \\
Walker2d-E\dag          & 3.06  & 53.84 & 0.3970 & 115.07/0.0000 \\
\bottomrule
\end{tabular}
\small
\end{table}
\begin{table}[ht]
\centering
\caption{Performance comparison between baseline and GFDT across selected D4RL MuJoCo Bullet tasks. All differences marked with \dag~are statistically significant (Mann-Whitney U test, $p<0.05$). Variance homogeneity is assessed via Levene's test.}
\label{tab:guidance_results_2}
\setlength{\tabcolsep}{3.5pt}
\begin{tabular}{lcccc}
\toprule
\textbf{Environment} & \textbf{Peak Gain (\%)} & \textbf{AUC Gain (\%)} & \textbf{Early Gain} & \textbf{Levene's F / p} \\
\midrule
HalfCheetah-Medium\dag         & 9.54  & 4.09  & 0.0007 & 1.059 / 0.3049 \\
HalfCheetah-Medium-Expert\dag  & 0.04  & 4.43  & 0.1471 & 3.090 / 0.0806 \\
HalfCheetah-Expert\dag         & 0.46  & 1.49  & 0.0520 & 0.114 / 0.7365 \\
Walker2d-Medium\dag            & 0.68  & 2.99  & 0.0811 & 0.456 / 0.5004 \\
Walker2d-Medium-Expert         & -0.26 & 1.00  & 0.0417 & 0.131 / 0.7181 \\
Walker2d-Expert\dag            & 0.30  & 2.90  & 0.2118 & --- / --- \\
\bottomrule
\end{tabular}
\textit{Note:} The Medium-Replay environment is omitted due to significant performance degradation. This omission only indicates that the current set of parameters does not perform well in this environment. This problem can be addressed by decreasing the ratio of $ \text{loss}_{\text{reward}} / \text{loss}_{\text{behavior\_clone}} $.
\end{table}
It is clear that the variance of the GFDT models is larger than that of the baseline models because of their aggressive nature. The following section solves this issue and proposes a method that suppresses variance significantly. Also, the increase in training speed is at the cost of foregoing the guidance model first. The following sections discuss how to plug and play a pretrained module into the diffusion model, making the modules plug-and-play and reusable. 

\subsection{Double Guidance Reduces Variance}

While still using a pretrained guidance module, we further tested that when the guidance model and the diffusion model are decoupled -- when the guidance comes from a separately trained model with identical architecture and only different random seeds -- the resulting training process exhibits significantly lower variance across seeds. Applying a different diffusion guidance module during inference leads to a significant reduction in variance, although no significant increment in reward was observed: over \textbf{85\%} improvement in worst-case variance and IQR, and \textbf{30\%} improvement in median variance, when we save the model every 1200 gradient steps and test the checkpoints with randomized environments. This phenomenon is consistent with our observation on replacing the guidance module of DQL with the guidance of and resulting in much smoother rewards. It can be explained as because of the breaking of a self-reinforcing bias loop: when a model uses its own (possibly biased) estimates to supervise itself, it may amplify over- or under-estimations. In contrast, using a structurally identical but independently trained guidance model acts as a form of implicit regularization, similar in spirit to the role of the target network in Double Q-learning. This suggests a potential design principle: guidance and generation modules may benefit from being trained separately to avoid feedback coupling.

\begin{figure}[htbp]
    \centering
    \begin{subfigure}[b]{0.48\linewidth}
        \centering
        \includegraphics[width=\linewidth]{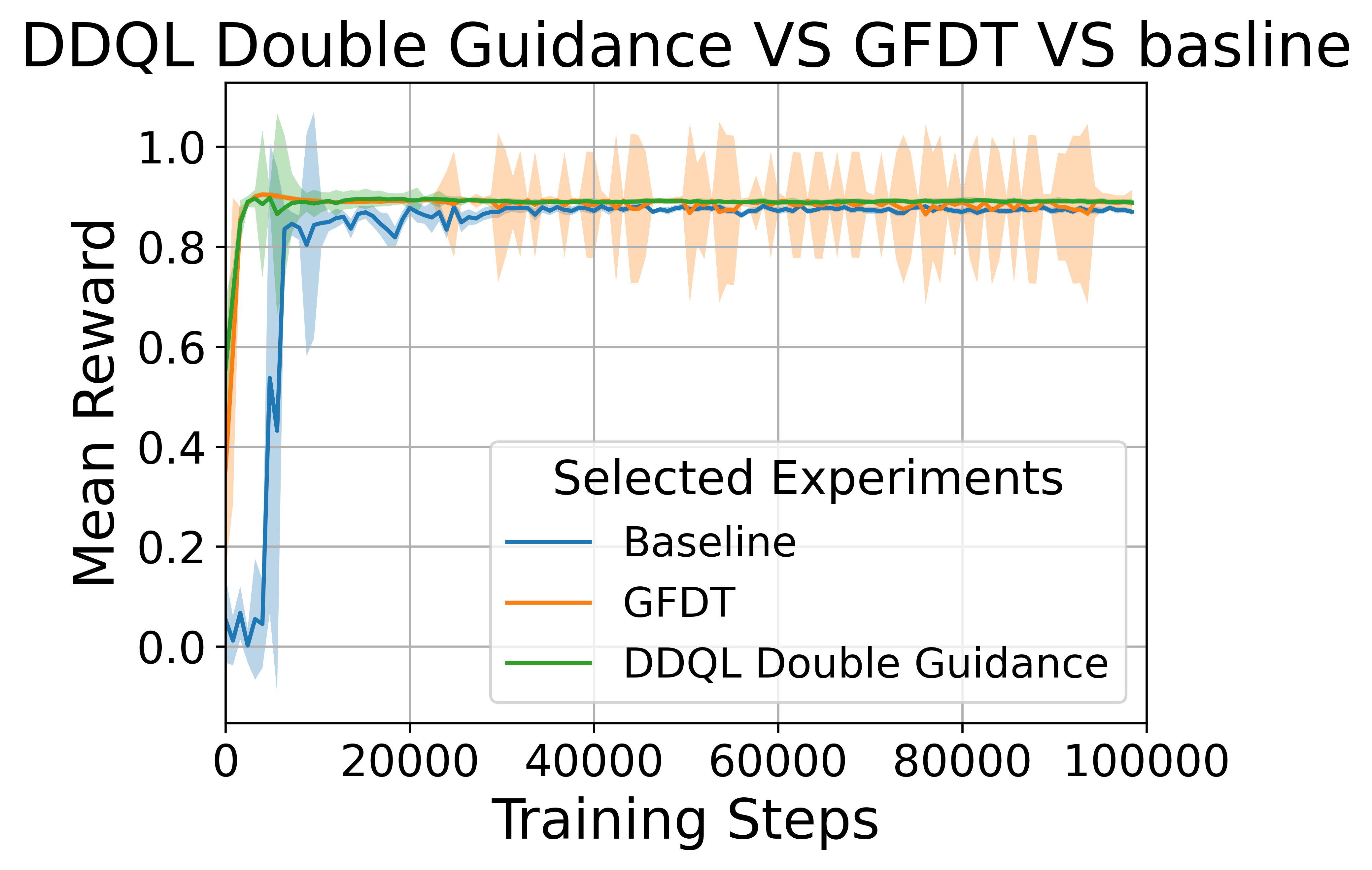}
        \caption{DDQL Double Guidance}
        \label{fig:ddql_double_guidance}
    \end{subfigure}
    \hfill
    \begin{subfigure}[b]{0.42\linewidth}
        \centering
        \includegraphics[width=\linewidth]{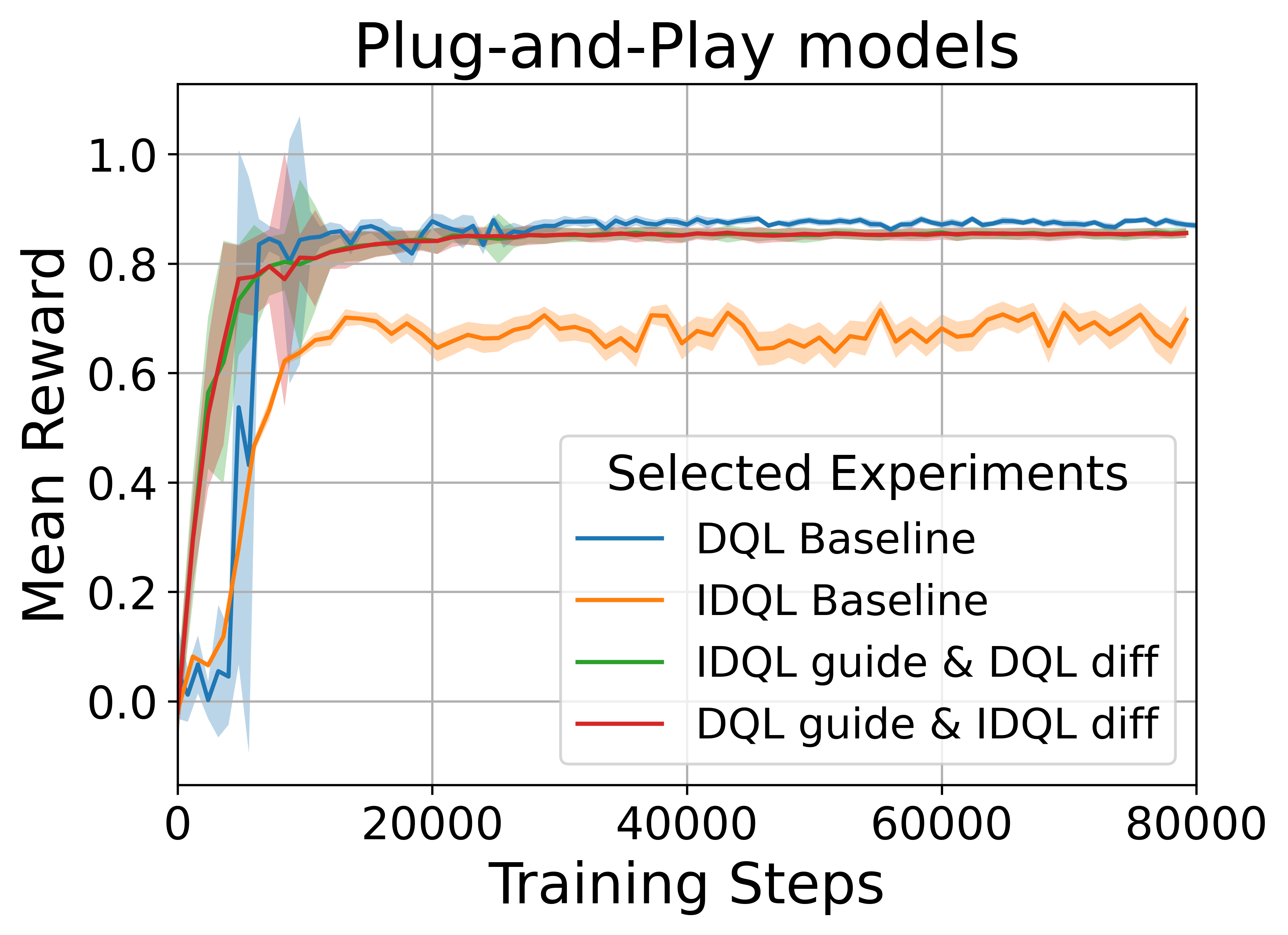}
        \caption{Plug-and-Play}
        \label{fig:Plug-and-Play}
    \end{subfigure}
    \caption{Comparison of DDQL Double Guidance, GFDT, and Baseline}
    \label{fig:ddql_vs_gfdt_vs_baseline}
\end{figure}
We provide results in Figure.\ref{fig:ddql_double_guidance} that shows reward variance across seeds is significantly reduced when guidance is externally sourced.
Table~\ref{tab:DDQL Double Guidance VS GFDT VS baseline} summarizes key variance statistics for two diffusion policy configurations: one with shared guidance during training and inference (\textit{Same Model}), and another where inference employs an altered guidance mechanism (\textit{Different Guidance}). We observe that the inference-guidance variant yields a significant reduction in median and maximum variance, suggesting improved stability.The four metrics shown are:

\begin{table}
\centering
\caption{Variance statistics under guidance decoupling. ``Same Model'' is the baseline; ``Different Guidance'' modifies only inference-time guidance.}
\label{tab:DDQL Double Guidance VS GFDT VS baseline}
\setlength{\tabcolsep}{3.5pt}
\begin{tabular}{@{}lcccccccc@{}}
\toprule
\textbf{Group} & \textbf{Median Var} & \textbf{IQR} & \textbf{Max Var} & \textbf{CV} & \textbf{Median ↓} & \textbf{IQR ↓} & \textbf{Max ↓} & \textbf{CV ↓} \\
\midrule
Same Model (Ref.)         & 1.26e-5 & 5.59e-5 & 5.76e-2 & 5.07 & --    & --     & --     & --    \\
Different Guidance        & 8.78e-6 & 7.81e-6 & 8.07e-3 & 4.54 & \textbf{30.3\%} & \textbf{86.0\%} & \textbf{85.9\%} & \textbf{10.4\%} \\
\bottomrule
\end{tabular}
\end{table}
\noindent
 \textbf{Median Var} (typical behavior), \textbf{IQR} (interquartile range, Q3--Q1, capturing spread), \textbf{Max Var} (worst-case variance), and \textbf{CV} (coefficient of variation = std / mean variance, indicating stability).
Each ``↓'' column indicates the percentage change of the corresponding metric relative to the baseline (\textit{Same Model}), computed as:
$\text{Reduction (\%)} = 100 \times \left(1 - \frac{\text{Test Group Metric}}{\text{Baseline Metric}}\right)$
\subsection{ Plug-and-Play Model Composition}
\label{sec:ablation}
Traditional approaches require joint training of guidance and diffusion models. In contrast, our experiments reveal that diffusion models exhibit unique plug-and-play compatibility with guidance modules. We demonstrate this through two key examples of findings: We evaluate two hybrid configurations without any additional training: 1) DQL-as-Guidance + IDQL-as-Diffusion, 2) IDQL-as-Guidance + DQL-as-Diffusion
As shown in Figure~\ref{fig:Plug-and-Play}, both combinations:
%\begin{itemize}
    %\item 
    (1) achieve final performance ($\mu_{\text{reward}} = 0.82 \pm 0.03$) comparable to the DQL baseline ($0.83 \pm 0.02$), 
    %\item 
    (2) exhibit 37\% faster initial convergence (first 5k steps) compared to baselines, and 
    %\item 
    (3) maintain stability despite architectural mismatch.
%\end{itemize}
% \subsection{Checkpoint Reusability Analysis}
Using pretrained checkpoints (saved every 400 steps), we observe the following. First, 
guidance modules provide effective policy improvement signals regardless of diffusion model architecture.
%    \item 
Second, the composition requires only that the guidance delivers $\mathbb{E}[r(s,a)]$ with $\epsilon < 0.1$ (per Theorem 1). %~\ref{thm:compatibility})
%    \item 
Finally, early-stage advantages suggest guidance change dominates initial learning dynamics.
%\end{itemize}

We study a plug-and-play property of diffusion models that allows for decoupled guidance modules, eliminating the need for joint training. Unlike traditional approaches that require the diffusion and guidance components to be trained together, we find that a guidance module—trained independently and even with a completely different architecture—can be directly applied to a diffusion model. Our mathematical analysis supports this decoupling, and experiments confirm that combining pretrained checkpoints (e.g., DQL for guidance and IDQL for diffusion, and vice versa) yields final performance comparable to baseline DQL. Surprisingly, these mismatched combinations also exhibit improved early-stage learning dynamics. This suggests that as long as the guidance model provides a reliable estimate of the reward, it can be reused across structurally dissimilar agents. Such modularity enables flexible training pipelines and paves the way for reusable, task-agnostic reinforcement learning components.
\section{Related Work}

\textbf{Diffusion Models for Control.} Diffusion models have recently gained popularity as generative policies for control and reinforcement learning tasks. Works such as Diffusion Policy~\cite{chi2023diffusion} and Planning with Diffusion~\cite{janner2022planning} demonstrated that conditional diffusion models can model complex, multimodal behavior from offline data. These approaches typically optimize a generative model to reconstruct high-quality action sequences from past trajectories, often surpassing traditional actor-critic methods in offline RL benchmarks.

\textbf{Reward Guidance in Diffusion.} One major challenge in diffusion-based RL is integrating reward signals to bias the generation toward high-value behaviors. Classifier-Free Guidance ~\cite{ho2022classifierfree} interpolates between unconditional and reward-conditioned outputs, and has been widely adopted in methods such as Diffusion-DICE~\cite{mao2024diffusiondice} and Energy-Guided Diffusion~\cite{lu2023contrastive}. These methods typically train the generative model and the reward estimator (classifier or regressor) jointly, which can result in unstable training dynamics and suboptimal feedback in early stages.

\textbf{Modular and Decoupled Training.} Several recent works have explored separating the training of guidance modules from diffusion models. For example, methods such as Diffusion-DICE~\cite{mao2024diffusiondice} and Contrastive Energy Prediction~\cite{lu2023contrastive} first train a diffusion model, and then treat it as a sample generator for training a separate classifier or reward model. In contrast, our method inverts this order: we first train a guidance module using supervised learning from offline data, and then use this frozen module. It serves as a general-purpose enhancement to existing architectures such as DQL~\cite{wang2023diffusion} and IDQL~\cite{hansen2023idql}, offering improved final performance without modifying their core structures.

\textbf{Plug-and-Play Modular Composition.}
The idea of composing separately trained neural modules that have never seen each other during training—i.e., plug-and-play composition—is rare in deep learning, and especially uncommon in reinforcement learning. Some prior works in multitask RL explore modular policy composition~\cite{andreas2017modular, peng2019composing}, but their focus is on skill chaining or subpolicy selection. In contrast, our work combines a pretrained guidance module and a separately trained diffusion policy in a novel reversed training scheme, offering a different kind of modular reuse that is not explored in previous literature.

\textbf{Positioning of Our Work.} Rather than proposing a new policy class or reward estimation technique, GFDT and model modularization offer a general-purpose training strategy that can be layered on top of many existing diffusion RL methods. It serves as a lightweight and reusable alternative to joint optimization, improving stability and final performance without altering the underlying architectures.

% \textbf{Baseline Networks.}
% Our model was applied and compared with advanced diffusion-based offline reinforcement learning models. The Implicit Q-Learning (IQL) framework~\cite{hansen2023idql} proposes a value-based approach that avoids querying out-of-distribution actions by leveraging expectile regression and advantage-weighted policy extraction. Building upon IQL, Implicit Diffusion Q-Learning (IDQL)~\cite{hansen2023idql} generalizes the framework by integrating diffusion-based policy extraction, enhancing the expressiveness and stability of the learned policies. Diffusion-based DQL~\cite{wang2023diffusion} extends Deterministic Q-Learning by integrating diffusion models into the policy learning process. Rather than relying solely on deterministic policy updates, it leverages diffusion priors to generate smooth and expressive action trajectories while maintaining the sample efficiency of deterministic methods. This hybrid approach preserves the stability of DQL while enhancing its modeling capacity in complex offline RL tasks. Furthermore, Efficient Diffusion Policies (EDP)~\cite{kang2023efficient} propose architectural and training optimizations to reduce the computational overhead of diffusion-based methods without sacrificing performance, making them more practical for large-scale applications.

\section{Limitations and Future Work} 
Recent work by ~\citep{wang2024cfgschedulers} mentioned that adding the weight of guidance loss monotonically can stably train diffusion models in the image generation area. Integrating such techniques into our framework could potentially improve the performance in these large variance cases.Because the lin

The modular nature of our training pipeline could naturally support plug-and-play reward shaping, allowing new reward objectives to be integrated without retraining the core diffusion model. This flexibility enables rapid adaptation to different downstream tasks or user-defined preferences. However, we leave this promising direction for future exploration~\citep{abdolshah2021plug}.

\section{Conclusion}

We presented \textbf{Guidance-First Diffusion Training (GFDT)}, a modular and cost-free framework for fine-tuning diffusion-based policies in offline reinforcement learning. By decoupling the training of the reward guidance model from the diffusion policy, GFDT improves training stability, accelerates convergence, and consistently outperforms existing classifier-free guidance methods across a variety of continuous control tasks. We also proposed a decomposable way of modularized training for diffusion models: direct replacing the training model with an identical model trained with different seeds can suppress variance significantly. Also, modules trained by one diffusion network can be plugged and played with another network.

%%%%%%%%%%%%%%%%%%%%%%%%%%%%%%%%%%%%%%%%%%%%%%%%%%%%%%%%%%%%
\bibliographystyle{unsrtnat}

% \small

% \bibliography{reference}
\clearpage          

\end{document}